%% file: main.tex
\documentclass[sigconf,table,screen]{article}

\usepackage[final]{neurips_2023}

\usepackage[utf8]{inputenc} 
\usepackage[T1]{fontenc}    
\usepackage{hyperref}       
\usepackage{url}            
\usepackage{booktabs}       
\usepackage{amsfonts}       
\usepackage{nicefrac}       
\usepackage{microtype}      
\usepackage{xcolor}         
\usepackage{graphicx}
\usepackage{subcaption}
\usepackage{enumitem}
\usepackage{algorithm}
\usepackage{algorithmicx}
\usepackage[noend]{algpseudocode}
\usepackage{amsmath}
\usepackage{tabularx}
\usepackage[symbol]{footmisc}

\newcolumntype{P}[1]{>{\centering\arraybackslash}p{#1}}
\newcolumntype{M}[1]{>{\centering\arraybackslash}m{#1}}

\title{Explainable Deep Multi-View Methodology\\for HEDP Foam Quality Assessment}

\author{Nadav Schneider \\
        Ben-Gurion University, IAEC \\
        Israel \\
  \texttt{nadavsch@post.bgu.ac.il} \\
  \And
   Muriel Tzdaka, Galit Strum, Guy Lazovski, \\ \textbf{Galit Bar, Gilad Oren, Raz Gvishi} \\
  SNRC \\
  Israel \\
  \texttt{rgvishi@soreq.gov.il} \\
  \And
  Gal Oren \\
  Technion, NRCN \\ 
  Israel \\
  \texttt{galoren@cs.technion.ac.il}
}

\begin{document}

\maketitle

\vspace{-0.5cm}
\begin{abstract}
Physical experiments often involve multiple imaging representations, such as X-ray scans, microscopic or spectroscopic images, and diffraction patterns. Deep learning models have been widely used for supervised analysis in these experiments. Combining these different image representations is frequently required to analyze and make a decision properly. Consequently, multi-view data has emerged -- datasets where each sample is described by multiple feature representations or views from different angles, sources, or modalities. These problems are addressed with the concept of multi-view learning. Understanding the decision-making process of deep learning models is essential for reliable and credible analysis. Hence, many explainability methods have been devised recently. Nonetheless, there is a lack of proper explainability in multi-view models, which are challenging to explain due to their architectures.

In this paper, we suggest four different multi-view architectures for the vision domain, each suited to another problem, with different relations between its views, and present a methodology for explaining these models. To demonstrate the effectiveness of our methodology, we focus on the domain of High Energy Density Physics (HEDP) experiments, where multiple imaging representations are used to assess the quality of foam samples. We expand the existing dataset and apply our methodology to classify the foam samples using the suggested multi-view architectures.
Through experimental results, we show an improvement by accurate architecture choice on both accuracy (78\% to 84\%) and AUC (83\% to 93\%) while presenting a trade-off between performance and explainability. Specifically, we demonstrate that our approach enables the explanation of individual one-view models through model-agnostic techniques, providing insights into the decision-making process of each view. This comprehensive understanding enhances the interpretability of the overall multi-view model.
The sources of this work are available at: \textcolor{blue}{\url{https://github.com/Scientific-Computing-Lab-NRCN/Multi-View-Explainability.git}}.
\end{abstract}


\maketitle

\section{Introduction}
\subsection{Multi-View Learning in Physics}
\label{multi_view_intro}
Physical experiments often involve multiple imaging representations, such as photographs, X-ray scans \cite{gales2004image}, microscopic or spectroscopic images of various types \cite{ge2020deep}, and diffraction patterns \cite{ferri2005high}. Deep learning models have been widely used for supervised analysis in these experiments \cite{de2019deep, das2022state, sadowski2018deep, rusanovsky2022end}. Combining these different image representations is essential for accurate classification and comprehensive analysis of complex phenomena \cite{sleeman2022multimodal, rusanovskyuniversal}. From high-energy physics experiments requiring data from different detectors \cite{schneider2022determining}, to astrophysics utilizing telescopes operating at various wavelengths \cite{reza2021galaxy}, multi-view data plays a critical role in classifying celestial objects. In material science, combining data from X-ray diffraction, electron microscopy, and spectroscopy helps characterize materials and study crystal structures \cite{ge2020deep, sumon2022multi, sumon2022deep}. Additionally, multi-view data is crucial in medical physics for precise disease diagnosis using imaging modalities like X-ray, MRI, CT, and ultrasound \cite{li2021deepamo, hosseinzadeh2022deep}. Fluid dynamics, biophysics, quantum optics, remote sensing, and other domains also benefit from multi-view data, enabling a deeper understanding of intricate physical processes and phenomena across different scales and disciplines. Consequently, multi-view data has emerged --- datasets where each sample is described by multiple feature representations or \textit{views} from different angles, sources, or modalities. These problems are addressed with the concept of \textit{multi-view learning} \cite{DBLP:journals/corr/LiYZ16, xu2013survey, sun2013survey, zhao2017multi, yan2021deep}.

Multi-view learning uses multiple feature sets or representations of data to improve the performance of machine learning algorithms. In the context of deep learning, there are several types of multi-view approaches, including multi-view Convolutional Neural Networks (CNNs), multi-view Auto-encoders, multi-view Generative Adversarial Networks (GANs), multi-view Graph Neural Networks (GNNs), multi-view Deep Belief Nets (DBNs), and multi-view Recurrent Neural Networks (RNNs), while each approach has its own unique advantage \cite{YAN2021106}.

Multi-view CNNs aim to learn high-level feature representations by integrating information from different views of the target data \cite{seeland2021multi}. To correctly classify these problems, the model has to consider each view and combine them into a joint decision. There are two main fusion strategies in multi-view CNNs: (1) the one-view-one-network strategy, where each view is processed by a separate CNN, and (2) the multi-view-one-network strategy, where multiple views are fed into the same CNN \cite{kan2016multi, seeland2021multi}. We are going to distinguish between different real-life tasks and propose different architectures based on these two strategies.

Despite the considerable success of deep learning-based multi-view approaches, they still encounter significant challenges. One such challenge is the requirement for abundant training data for each view to effectively capture diverse representations and correlations, which can be both time-consuming and costly \cite{zhao2017multi}. Additionally, a major limitation lies in the lack of explanations for the decisions made by the complex architectures of deep learning models \cite{yan2021deep, joshi2021review, rahate2022multimodal, saeed2023explainable}, presenting a considerable obstacle in critical applications where interpretability is crucial\footnote{We note that interpretable models can be easily understood by humans without any additional aids or techniques, as their decision-making process is transparent and straightforward. On the other hand, explainable models do provide explanations for their predictions, but a deeper understanding may require supplementary tools or techniques, as their internal mechanics may not be immediately apparent to humans. In the context of this work, the two terms are applicable and interchangeable in different contexts, as we are trying to interpret the multi-view model decisions in the experiments while we also present an explainability methodology to allow such interpretability.}. 

\subsection{Multi-view Model Explainability}
Explainability in the context of deep learning is the research field of explaining deep learning models' decisions. An ability which is often crucial in high credibility obligated systems, like autonomous driving \cite{zablocki2022explainability}, medical imaging \cite{SHACHOR20201, Multi_view_breast} and physics-oriented problems \cite{roscher2020explainable} in particular. Explaining deep learning models' decisions is a growing research field, and these models are no longer considered as a complete "black box" \cite{explainable_deep, hödl2023explainability}. 

Nonetheless, the explainability of multi-view models remained mostly unaddressed. This is due to the challenges caused inherently by these architectures and will be discussed in \autoref{multi-view_exp}. While many explanation methods are proposed \cite{joshi2021review}, and some even applied for multi-modals \cite{DBLP:journals/corr/abs-1802-08129}, they are limited and are not suited for any multi-view model, and specifically, to multi-view CNNs in the vision domain, especially when particular classification should be made. In experimental physics-oriented problems, multiple image representations are extracted through an experiment, and a certain classification should be made (\autoref{multi_view_intro}). Explainability is also critical to ensure reliable and credible analysis.

Thus, current explanations in computer vision multi-view classification models mostly rely on the two common types of explainers that are able to achieve local (such as Local interpretable Model-Agnostic Explanations (LIME) \cite{lime}) or global (such as SHapley Additive exPlanations (SHAP) \cite{SHAP}) interpretability for each specific view \textit{separately} (while disregarding the actual "big picture" the multi-view model was designed to provide initially).
Note that both LIME and SHAP algorithms are model-agnostic algorithms, meaning they can be applied to any machine learning model, regardless of its complexity or architecture. This is because they only consider the input examples and their corresponding predictions.

An explanation may appear in various options in the different domains. In computer vision tasks, the input image pixels are marked with relevant areas for the model. These marked areas can be represented as contributing areas for a specific class where each class has its own color and even as a heat map where a gradient of colors represents the model's focus during its prediction. These types of explanations are common both in classification and segmentation tasks. This allows users to visually understand which regions of the image are influencing the model's prediction for a particular class. These color-coded areas provide either local or global interpretability (again, per view), offering insights into the decision-making process for individual instances \cite{schorr2021neuroscope}. However, when dealing with multi-view models, this approach becomes problematic. In multi-view models, the input comprises multiple feature representations or "views" from different sources or modalities. The challenge lies in combining these views into a single decision, which often involves a view-pooling step where features are blended, leading to the loss of individual view-specific information. As a result, the model's attention or focus on specific areas in each view is not easily discernible, and explanations may not accurately reflect the true contribution of each view to the final prediction. Thus, achieving explainability in multi-view models remains a challenging task due to their inherent complexities \cite{roscher2020explainable}.

\subsection{Contribution}

In this paper, we suggest four different multi-view CNN architectures for the vision domain, each suited to another problem, with different relations between its views, and present a relatively simple methodology for explaining these models and interpreting the decision-making process for experimental physics. Then we apply the methodology fully to a real-world physics experiment problem of High Energy Density Physics foam quality classification and generate several explanations (\autoref{fig:teaser}). The main contributions of this paper are as follows:
\begin{enumerate}[wide, labelwidth=!, labelindent=0pt]
\item Proposal of four novel multi-view architectures for the vision domain, each suited to specific problem types and relations between views, leading to improved performance and interpretability in physical experiments and their decision-making processes.
\item Comprehensive trade-off analysis between performance and explainability, offering insights into the strengths and limitations of each multi-view architecture and guiding future model design and optimization.
\item Demonstrating the proposed architectures and methodology in a real-world physics experiment involving High Energy Density Physics (HEDP) foam quality classification, showcasing significant accuracy and AUC improvements compared to previous work.
\item Potential applicability of the proposed methodology and architectures to other scientific domains beyond HEDP experiments, expanding the scope of multi-view model applications in various fields.
\end{enumerate}

\input{figures_0}
\input{figures_1}
\section{Multi-View Architectures}
\label{multi_view_architectures}
We begin by presenting and defining four different relevant multi-view architectures, each assuming various relations between the views and, as such, fits another type of multi-view explainability (\autoref{fig:multi_view_architectures}). We observe that multi-view architecture suitability to the relevant needed explainability type directly correlates to the overall performance (see \autoref{results}), as the model   understands the influence of the diverse features differently in each architecture (see \autoref{multi-view_exp}):
\begin{enumerate}[wide, labelwidth=!, labelindent=0pt]
    \item \textbf{Completely Similar Views (CSV)} --- This approach assumes the different views are visually similar. Therefore, one feature extractor applies to all of the views under the assumption that it could efficiently extract relevant features from all of them. A pooling view through all of the outputs occurs, and the resulting vector is connected to a classifier. This architecture is mostly useful when there are multi-views of one object in one form (\autoref{mv4}).
    \item \textbf{Similar Sub-Groups (SSG)} --- This approach assumes clustering views to two or more sub-groups based on similar visual attributes is possible. Each group has its own feature extractor, and therefore, each feature extractor is optimized for its specific sub-group. Each sub-group can be visually different and still preserves an optimal extraction for each (\autoref{mv2}). 
    \item \textbf{Partly Similar Groups (PSG)} --- This approach assumes clustering views to two or more sub-groups based on similar visual attributes is possible. Nonetheless, views in sub-groups still have their unique features. Hence, each view has its own feature extractor. Then, each sub-group is passed through a view-pooling layer and a classifier, leading to a complete separation from other sub-groups during the classification process. Finally, each sub-group outputs a softmax decision to an additional classifier which then outputs the final decision (\autoref{mv1}).
    \item \textbf{Completely Different Views (CDV)} --- This approach assumes each view has its own unique visual appearance, and thus, one feature extractor could not learn to extract relevant features from all views. Therefore, a feature extractor is applied for each view, then view pooling and a classifier (\autoref{mv3}).
\end{enumerate}

\section{Multi-View Explainability}
\label{multi-view_exp}
\subsection{Problem Definition}
Given the multi-view architectures, defined in \autoref{multi_view_architectures}, we move to the needed explainability. Multi-view models' explainability is not trivial due to feature maps loss of their corresponding view after the feature map combination (view-pooling) process. A mixture of these feature maps is necessary to exploit all the data effectively and to reach maximal performance (\autoref{multi_view_intro}). Consequently, the multi-view models' explanation is inherently limited. Providing effective and comprehensive explainability for general and the suggested multi-view architectures (which are relevant in the context of physics experiments, \autoref{multi_view_intro}) is the problem we want to address.

\subsection{Suggested Solution}
We propose a methodology for explaining multi-view models given the different multi-view architectures (\autoref{fig:multi_view_architectures}). To overcome the view-pooling, in which afterward, features are blended and contribution from each view is lost, we propose the following steps (also presented formally in Algorithm \ref{algo1}): (1) Train the feature extractor inside the multi-view mechanism, (2) Freeze its weights and attach the trained feature extractor for each view, (3) Attach a classifier layer at the end of each view, (4) Train the resulting one-view models, and (5) Activate an explanation algorithm for each one-view model.


Steps 2-4 are represented by EXPLAINer in \autoref{fig:multi_view_architectures} and illustrated in-depth in \autoref{fig:explainer}. To explain each view for each multi-view architecture properly, slight modifications were performed:
\begin{enumerate}[wide, labelwidth=!, labelindent=0pt]
    \item \textbf{CSV} --- One feature extractor learns to extract relevant features in a relatively accurate way. Hence, the frozen feature extractor is attached to all views, and then there is one EXPLAINer box to all of the views (\autoref{mv4}).
    \item \textbf{SSG} --- Each sub-group has one feature extractor and, therefore, will be concatenated to one EXPLAINer (\autoref{mv2}).
    \item \textbf{PSG} --- Views in sub-groups are visually similar but also different in a way; therefore, each view in a sub-group has its own feature extractor, and then each feature extractor is concatenated to an EXPLAINer (\autoref{mv1}).
    \item \textbf{CDV} --- Each view has its own EXPLAINer box since each view has its own unique feature extractor (\autoref{mv3}).
\end{enumerate}

\input{figures_2}

The proposed methodology enables relevant feature extraction from each view, sub-group, or group, depending on the applied multi-view architecture. The resulting explainability is the model's attention, meaning each marked area contributes to one of the classes, but we can not decide whom to which. By that, and contrary to general purpose multi-view classification, which does not use the devised architectures, it is possible to give an accurate explainability (either local or global) per EXPLAINer and interpretability of the entire multi-view classification decision-making visually. EXPLAINer output is a classification but is only used through model agnostic techniques to mark the contributing areas in the images. This is by extracting the features treasured inside the feature extractors (CNNs in our case).

We note that there is a trade-off between the explainability capacity of the suggested architectures and their performance. This is due to the fine-tuned one-view classifier, which trains based on a frozen feature extractor on one of the views. This training inherently limits the model's performance. A classifier can not properly classify when it has not been trained on the other views as well. Nonetheless, the explainability is fine-tuned to the specific view, and connections are built between the frozen feature extractor and the corresponding relevant features in the view, which is then emphasized. In the following use case of HEDP experiments (\autoref{HEDP}, we will elaborate on this trade-off in practice.



\section{High Energy Density Physics Use Case}
\label{HEDP}
To demonstrate our methodology, we aimed to enhance both performance and explainability compared to previous research in this field by \cite{schneider2022determining}. In that work, a real-world problem has been addressed in the domain of High Energy Density Physics (HEDP) experiments \cite{ref1, ref2, ref3, ref4, harel2020complete}. HEDP involves a dynamic wave-front propagating inside a low-density foam, which is commonly aerogel. Aerogels are a large family of materials, generally defined as extremely low-density solids (more than 90\% porosity, less than 200mg/cm$^3$ density) \cite{ref7, ref8}.

A significant limitation of these aerogels is their tendency to suffer from cracks \cite{ref32, ref33, ref34}. Therefore, it is required to characterize each sample for its quality as a foam. A single aerogel specimen includes five images (\autoref{fig:teaser}). An expert searches for characteristics such as scratches, dirt, and dark stains in the top and bottom view images acquired by an optical microscope, implying a deep hole inside the foam. As a complementary, the profile images -- acquired by laser scanning confocal microscopy (LSCM) -- help confirm or refute the initial assumptions. This complicated, harsh, and exhausting work of manually classifying the foams' quality through the image set is mandatory to decide whether foams can be used in experiments (such as HEDP). Expert labels the quality levels to either Normal (Valid to use in experiments), Defective (Not valid), or Normal-Defective (contains defective characteristics, yet they are not sufficient to determine a defective classification). 

The authors in \cite{schneider2022determining} present a multi-view deep-learning solution for the specified problem. However, explainability for the multi-view model is absent, and its architecture is not suited to the given task. The authors use Completely Similar Views (CSV) architecture and several other one-view model configurations are suggested. Input alters accordingly: top and bottom views only, profiles only, and full group (all-views). The given explanation is partial and limited to one-view models only. An issue was presented with the use of normal-defective examples. On the one hand, due to the limited number of examples, ignoring them is not ideal. On the other hand, these examples might confuse the models. Finally, models were trained both with and without normal-defective examples.

In our work, and for clarity, we expand the previous dataset (\autoref{dataset}) and exclude normal-defective examples, focusing on the all-views configuration. We also demonstrate how different multi-view architecture affects learning, and how better results and interpretability are achieved when using the proper architecture. The suggested architectures applied for the given use case are presented in \autoref{fig:multi_view_demonstrate}.

\input{figures_3}

\subsection{Dataset}
\label{dataset}
As previously mentioned, additional data has been added, turning the data set to a total of 123 labeled examples, each containing 5 views (615 images in total). This is in contrast to the previous work, with only 95 labeled examples. The data set was originally created for an expert's usage and not for a learning model. Repeatable methods for adequate image acquisition do not exist, this leads to a large variance between examples, and therefore, to adjust the data set into a learning model, a massive pre-processing is required. We are using the same pre-processing made in \cite{schneider2022determining}, and the only change is data set expansion.

\input{figures_5}

\subsection{Experimental Results}
\label{results}
Since more data has been made available, normal-defective examples, which might confuse the model, have been removed. For a valid comparison, the previous model --- CSV multi-view without normal-defective examples model has been retrained on the expanded data set.
The data set split is a 70:30 train-to-test ratio. Overall, without the normal-defective examples, there are 59 training examples, 29 normal and 33 defectives, and 19 test examples --- 10 normal and 9 defectives.

The different multi-view models were trained and executed on a 32GB Tesla V100 GPU, using Pytorch \cite{paszke2019pytorch}. The training was carried out with a batch size of 4, 150 epochs, a cross-entropy loss function, a learning rate of 0.00005, and an Adam optimizer. Accuracy and AUC were measured with the predefined settings for the different architectures. Loss and accuracy trends of each multi-view model and their performance are presented in \autoref{table:learning_curve}.

\subsubsection{Evaluation}
In \cite{schneider2022determining}, all-views model without normal-defective examples and yielded on the test set accuracy of 78\% and AUC of 75\%. After re-training with the same settings and new additional data, the accuracy remained at 78\% while its AUC raised to 83\%.
We show that the SSG architecture is best suited for the given problem and therefore achieves the best accuracy -- 84\%, and AUC -- 93\%. This is compared to the previous work that used a non-optimal architecture (CSV) for the given problem. Loss and accuracy trends of the new architecture converge better and indicate better learning (\autoref{table:learning_curve}). Note that AUC indicates the model's generalization capability and constitutes a crucial metric due to the tiny number of examples in the test set.

\subsubsection{Explainability}
We train the one-view classifiers according to Algorithm \ref{algo1} and generate an explanation for each view based on the most relevant features. These models can be used to explain the prediction both locally (\autoref{fig:LIME_attention}) and globally (\autoref{fig:SHAP}), in a way that collectively the experiment multi-view deep-learning classification model turns interpretable for further decision-making. As mentioned, the explanation is limited to merely attention areas since the one-view models can not correctly classify based on one-view only. There is a trade-off between performance and explainability; SSG leads to the highest results but there is one CNN for multiple views, and therefore, explainability per view is damaged. CDV on the contrary has CNN for each view, therefore, explainability will be better, however, the performance is not optimal.

\input{figures_4}

\section{Conclusions and Future Work}
In this paper, we proposed a novel approach for explainability in multi-view deep learning models applied in the domain of High Energy Density Physics (HEDP) experiments with foam samples. We introduced four multi-view architectures: Completely Similar Views (CSV), Similar Sub-Groups (SSG), Partly Similar Groups (PSG), and Completely Different Views (CDV), tailored for different physics-oriented problems. Our methodology involved training one-view classifiers with frozen feature extractors for each view, enabling effective feature extraction and interpretation. The SSG architecture showed the best performance, achieving an accuracy of 84\% and an AUC of 93\% in classifying foam samples. 

For future work, we believe that extending the methodology and architectures to other scientific domains, such as medical imaging and materials science, will validate the approach's efficiency in various contexts. Exploring hybrid architectures that combine multiple multi-view approaches offers the potential for more robust and effective models, capitalizing on the strengths of different techniques. Lastly, incorporating human-in-the-loop approaches leverages domain expertise, enhancing overall model performance and trust. Embracing these directions will promote the field of explainable multi-view models and foster their broader adoption in various scientific disciplines.

\subsubsection*{Acknowledgments}
This research was supported by the Pazy Foundation and the Lynn and William Frankel Center for Computer Science. Computational support was provided by the NegevHPC project~\cite{negevhpc}. The authors would like to thank Hannah Zilverberg, Matan Rusanovsky, Israel Hen, and Gabi Dadush for their technical support.

\bibliography{iclr2024_workshop}
\bibliographystyle{iclr2024_workshop}

\end{document}

%% file: figures_0.tex
\begin{figure*}
\centering
  \includegraphics[width=1\textwidth]{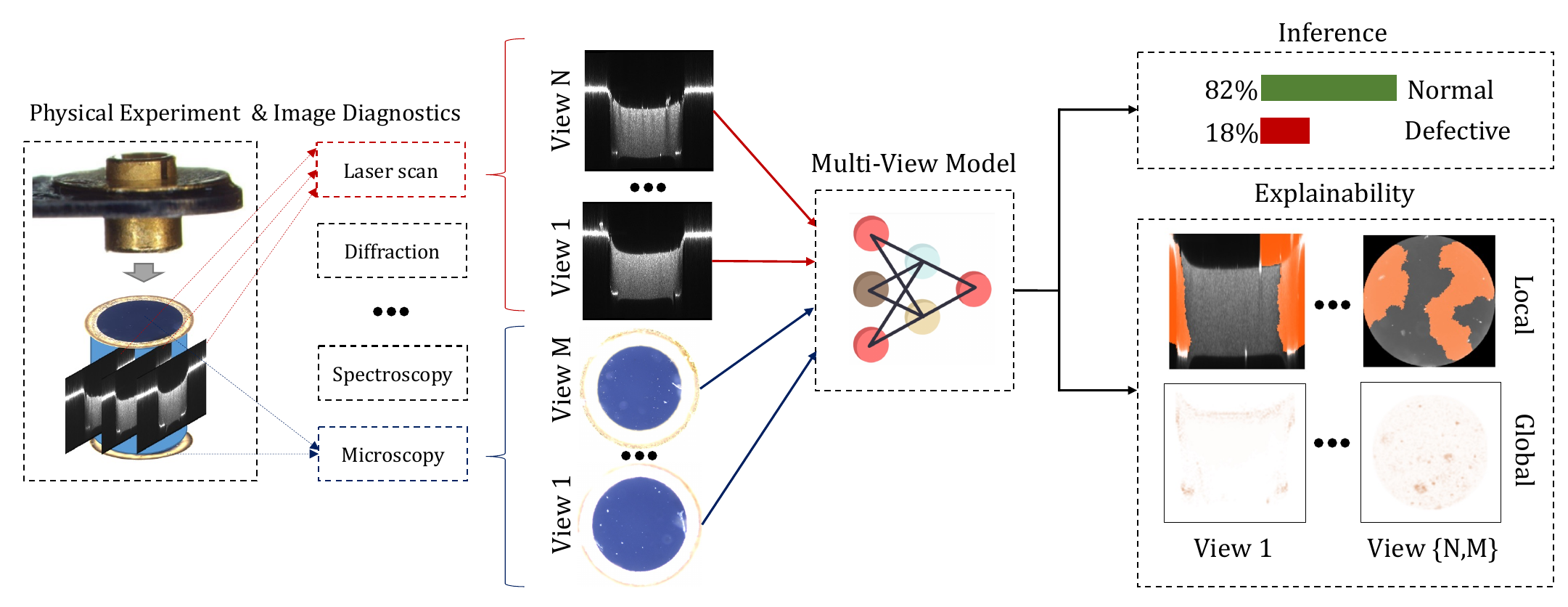}
  \vspace{-0.8cm}
  \caption{Multi-View model gets as an input M+N different views from a certain image diagnostic. In this example, foam quality is assessed for HEDP experiments. There are two visually different sub-group views. Then, the model outputs a classification and the explanation behind it, explainability is both local and global.}
  \label{fig:teaser}
\end{figure*}

%% file: figures_1.tex
\begin{figure*}[!ht]
  \centering
  \captionsetup[subfigure]{}
  \subcaptionbox{Completely Similar Views\label{mv4}}{\includegraphics[width=1.2in]{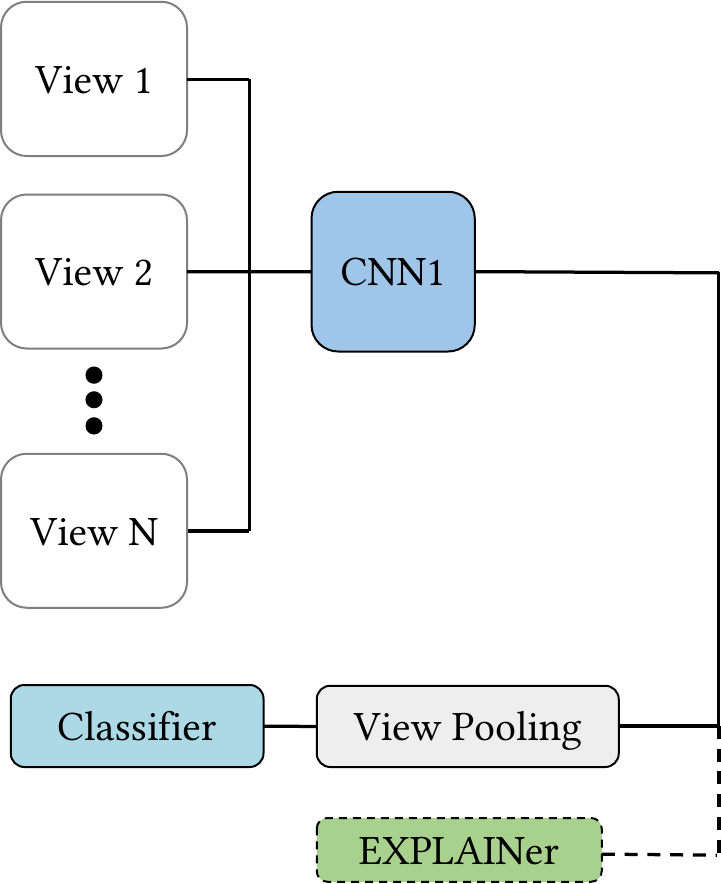}}\hspace{1em}
    \subcaptionbox{Similar Sub-Groups\label{mv2}}{\includegraphics[width=1.2in]{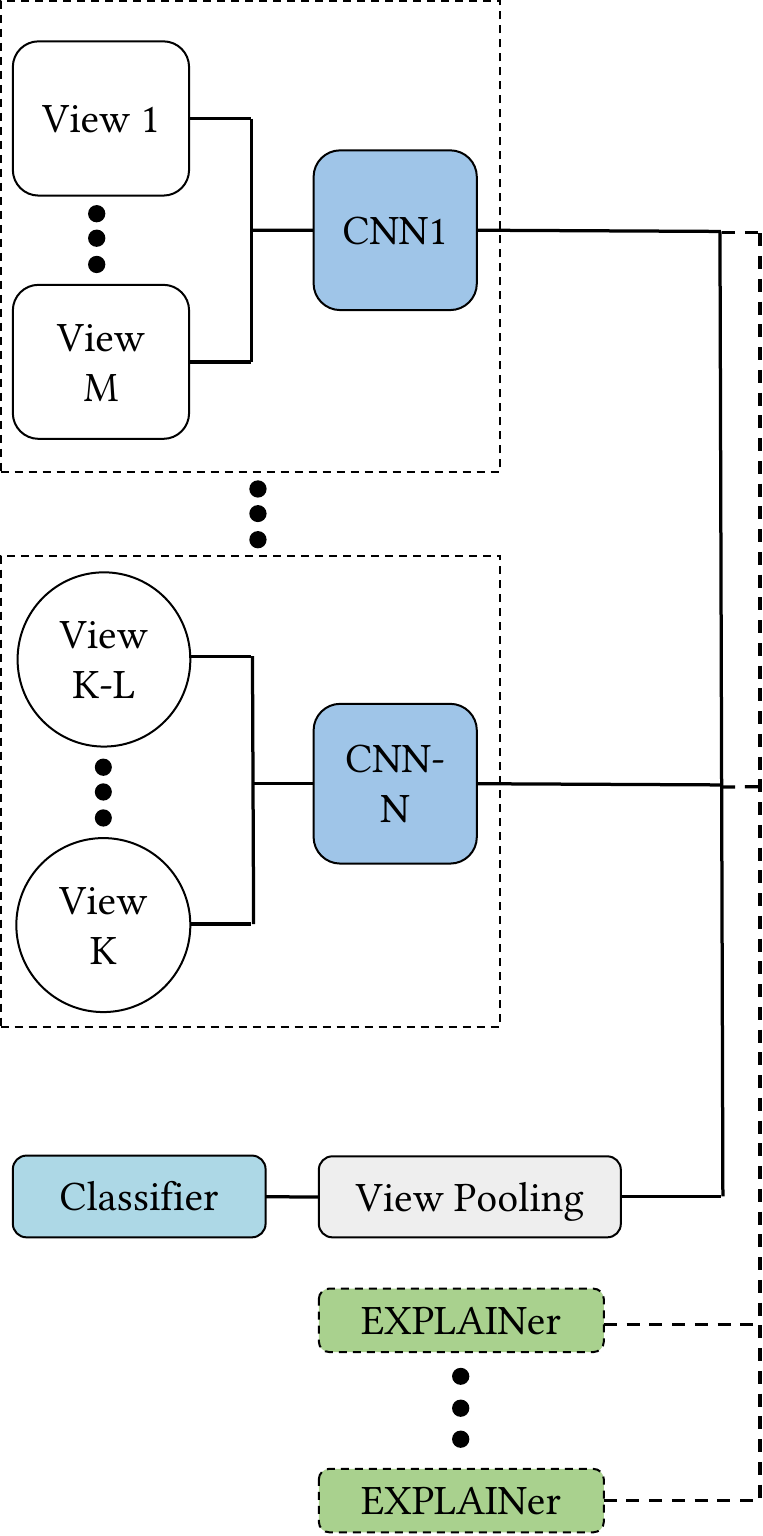}}\hspace{1em}
   \subcaptionbox{Partly Similar Groups\label{mv1}}{\includegraphics[width=1.2in]{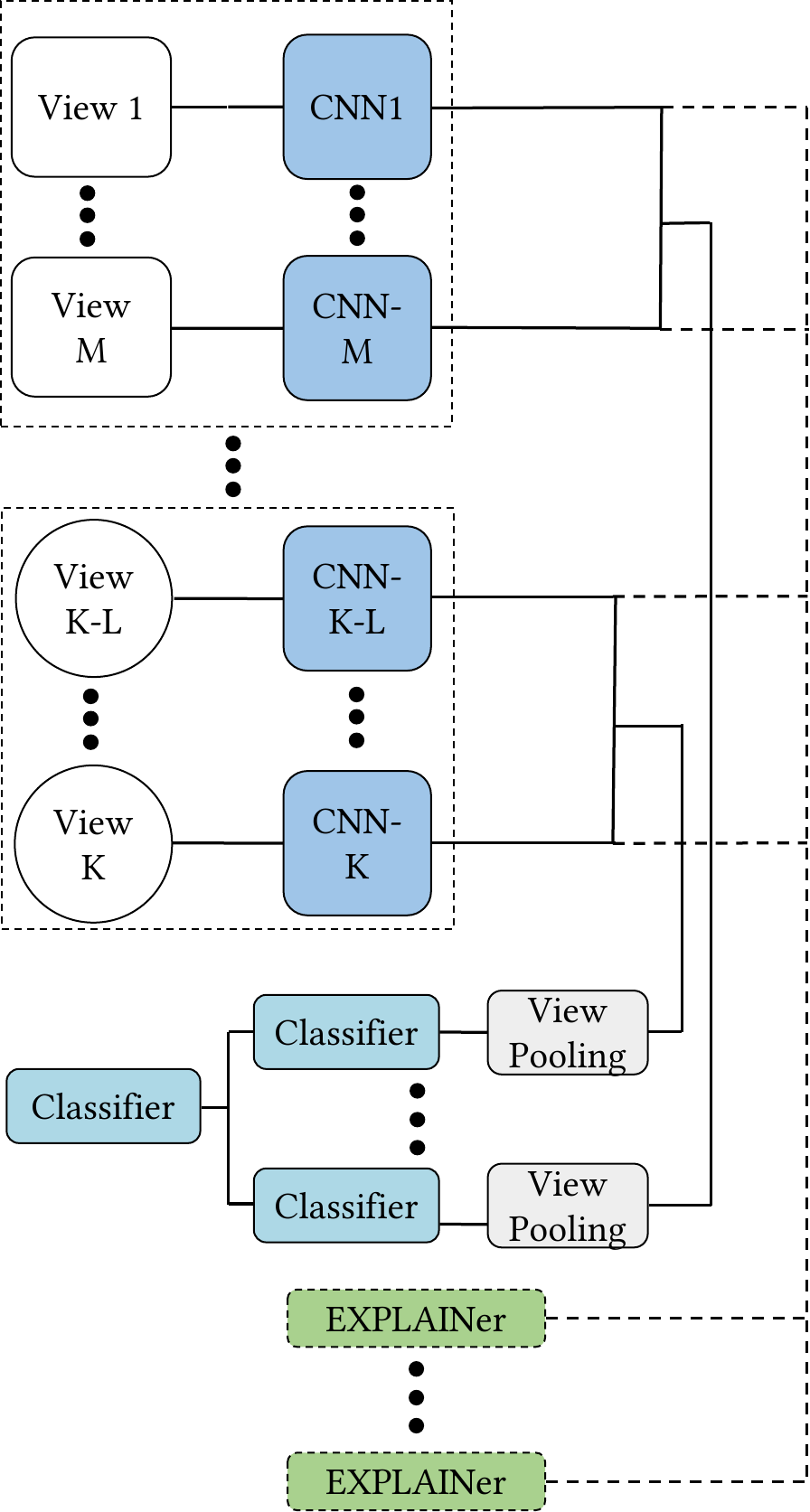}}\hspace{1em}\vspace{1.5em}
  \subcaptionbox{Completely Different Views\label{mv3}}{\includegraphics[width=1.2in]
  {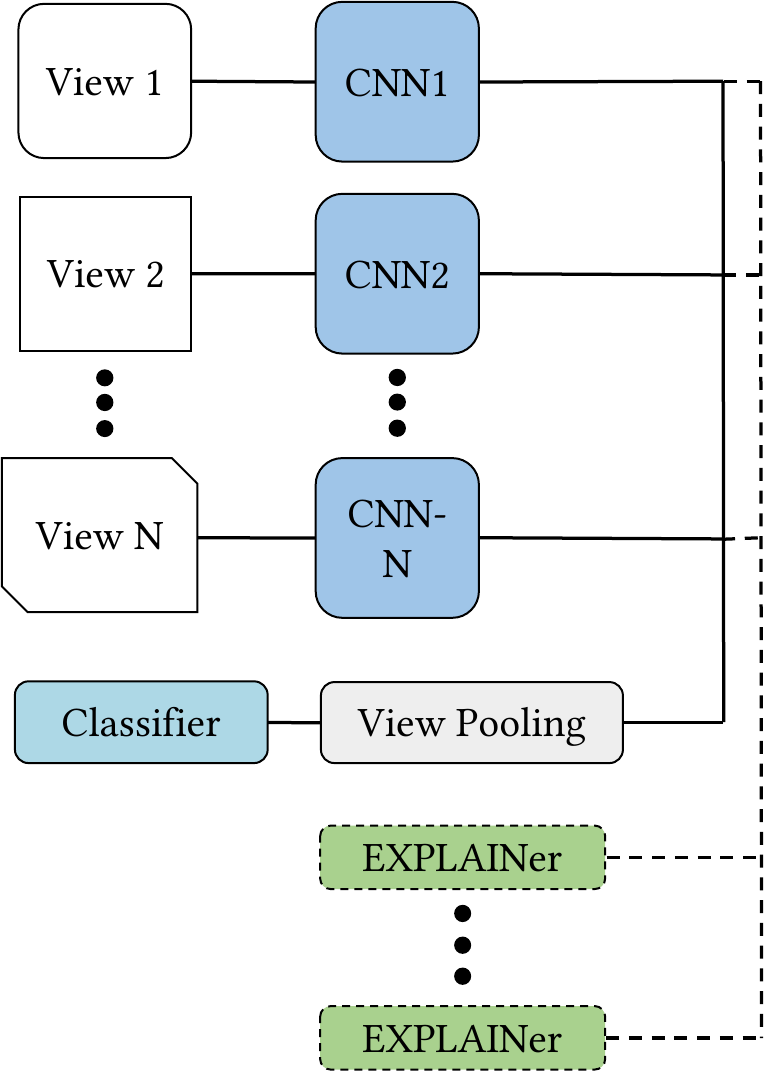}}\hspace{1em}
  \vspace{-0.8cm}
  \caption{The four different suggested multi-view architectures which are suited to different problems: (a) assumes views are similar, (b) assumes visually similar sub-groups, (c) assumes similar sub-groups but also different in a way, and (d) assumes views are completely different. EXPLAINer content is demonstrated in \autoref{fig:explainer}. Note that CNN can be replaced with any other feature extractor.}
  \label{fig:multi_view_architectures}
\end{figure*}

%% file: figures_2.tex




\begin{figure}[!ht]
\centering
\begin{minipage}{0.55\textwidth}
\begin{algorithm}[H]
\caption{Multi-View Explainability}
\label{algo1}
\begin{algorithmic}[1]
\Procedure{MVExplain}{$\text{CNNs, classifiers}$}
    \For{$\text{CNN in } \text{CNNs}$}
        \State $\text{CNN}.\text{freeze}()$ 
        \State $\text{input} \gets \text{Empty list}$

        \For{$\text{view in CNNs[i].input}$}
            \State $\text{cnn\_output} \gets \text{CNNs}[i].\text{forward}(\text{view})$ 
            \State $\text{input.append}(\text{cnn\_output})$ 
        \EndFor

        \State $\text{classifiers}[i].\text{fit}(\text{input})$ 
    \EndFor

    \For{$\text{classifier in classifiers}$}
        \State $\text{Apply LIME to }\text{classifier}$ 
        \State $\text{Apply SHAP to }\text{classifier}$ 
    \EndFor
\EndProcedure
\end{algorithmic}
\end{algorithm}
\end{minipage}%
\hfill
\begin{minipage}{0.4\textwidth}
    \centering
    \includegraphics[width=\textwidth]{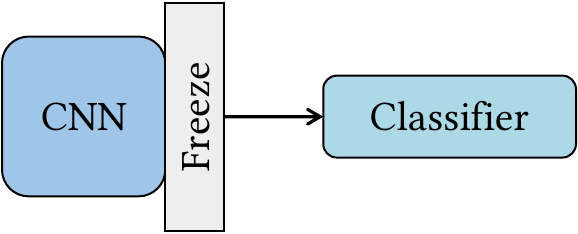}
    \vspace{-0.3cm}
    \caption{This scheme defined as EXPLAINer in \autoref{fig:multi_view_architectures} and \autoref{fig:multi_view_demonstrate}. EXPLAINer action is freezing the concatenated feature extractor weights and training a classifier (Algorithm \ref{algo1}).}
    \label{fig:explainer}
\end{minipage}
\end{figure}

%% file: figures_3.tex
\begin{figure*}
  \centering
  \captionsetup[subfigure]{}

    \subcaptionbox{Completely Similar Views\label{mv4_true}}{\includegraphics[width=1.2in]{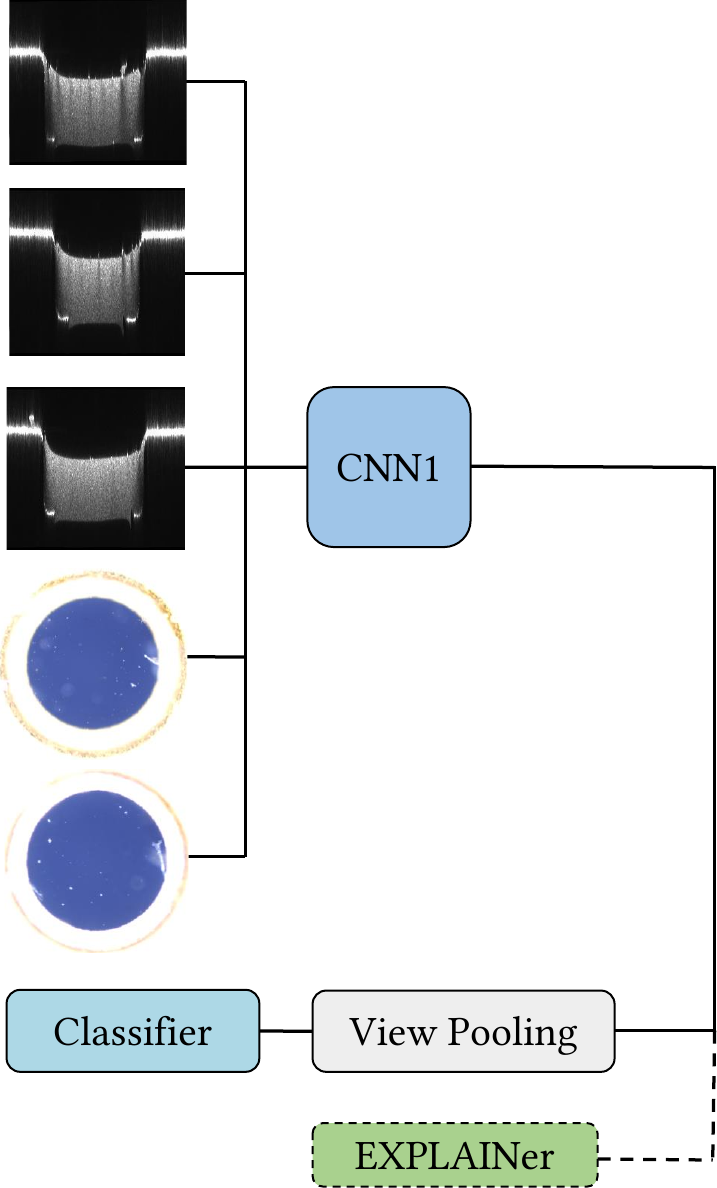}}\hspace{1em}%
  \subcaptionbox{Similar Sub-Groups\label{mv1_true}}{\includegraphics[width=1.2in]{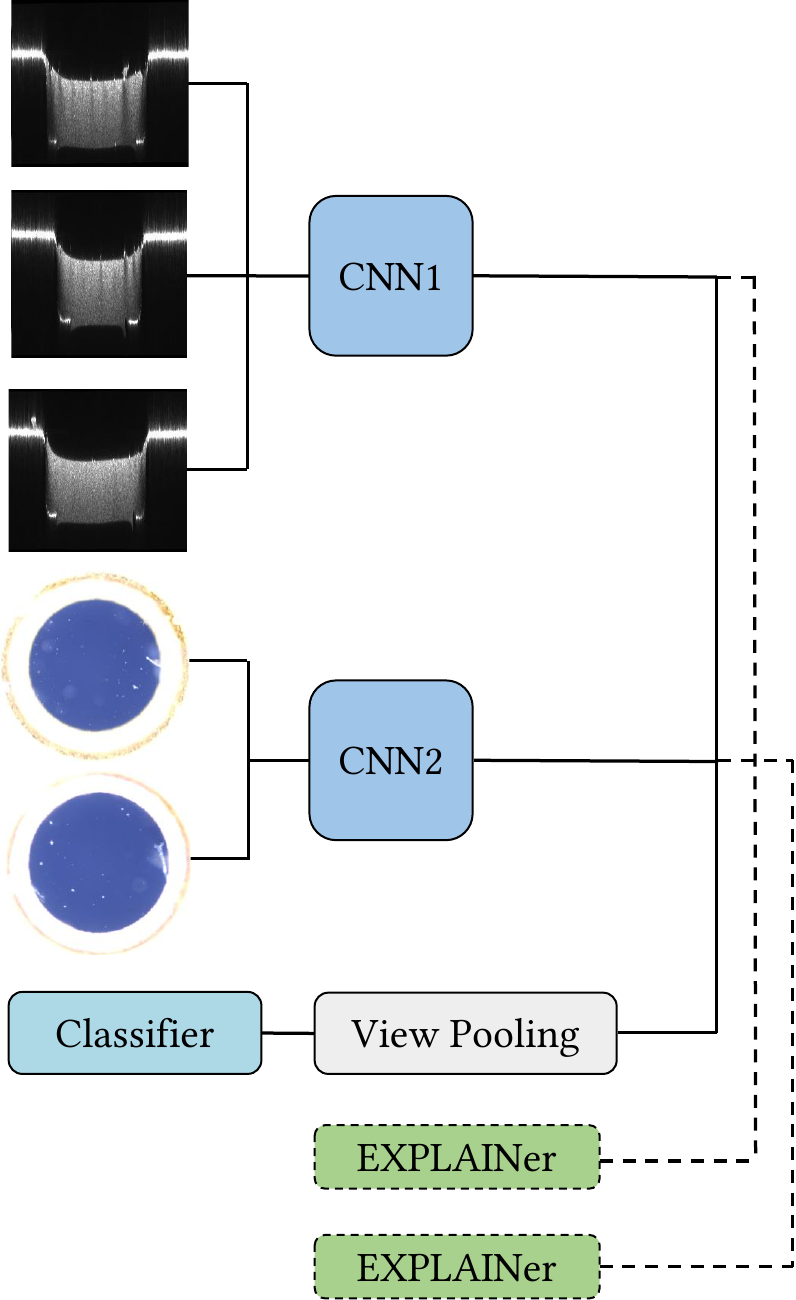}}\hspace{1em}\vspace{1.5em}%
  \subcaptionbox{Partly Similar Groups\label{mv2_true}}{\includegraphics[width=1.2in]{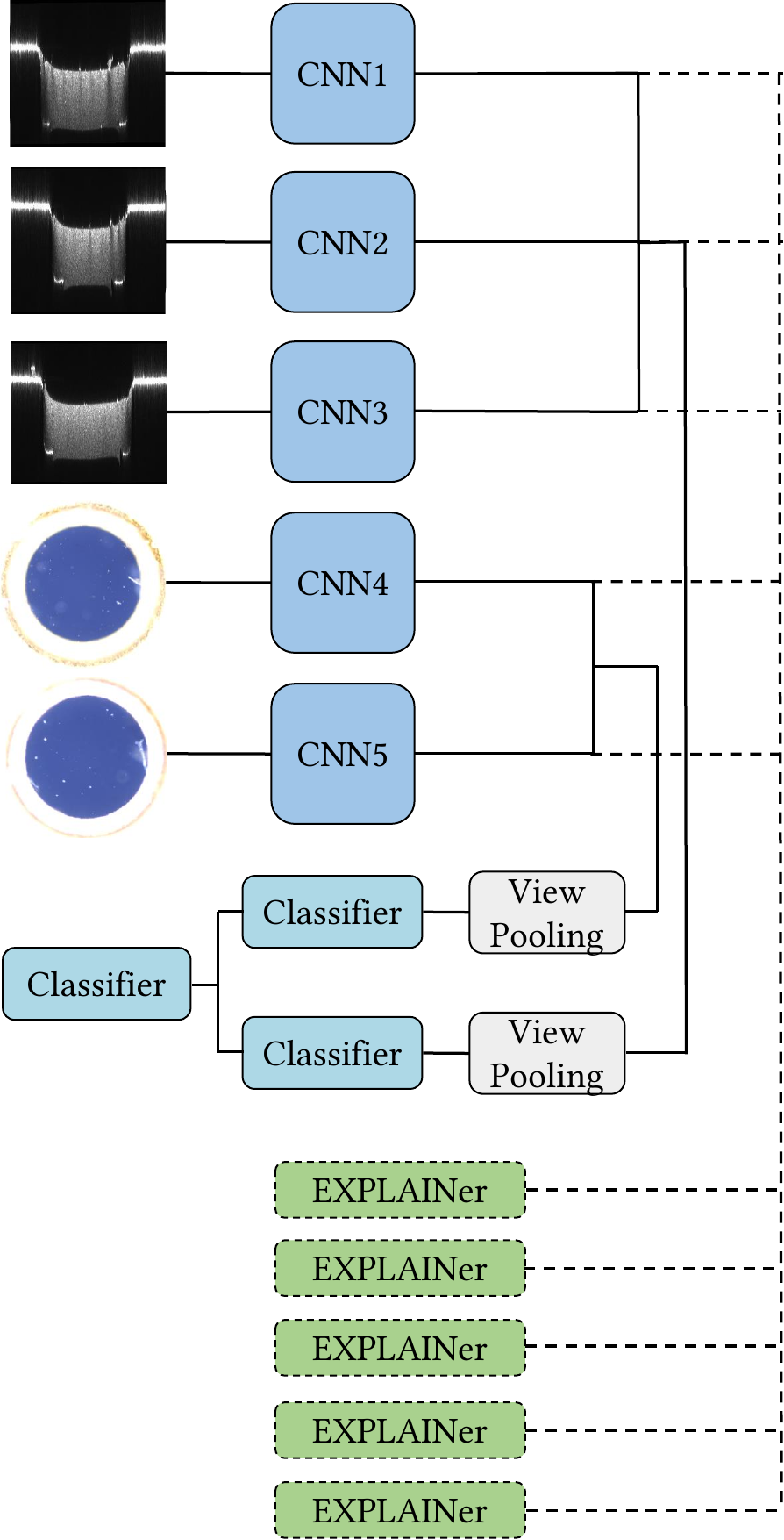}}\hspace{1em}%
  \subcaptionbox{Completely Different Views\label{mv3_true}}{\includegraphics[width=1.2in]
  {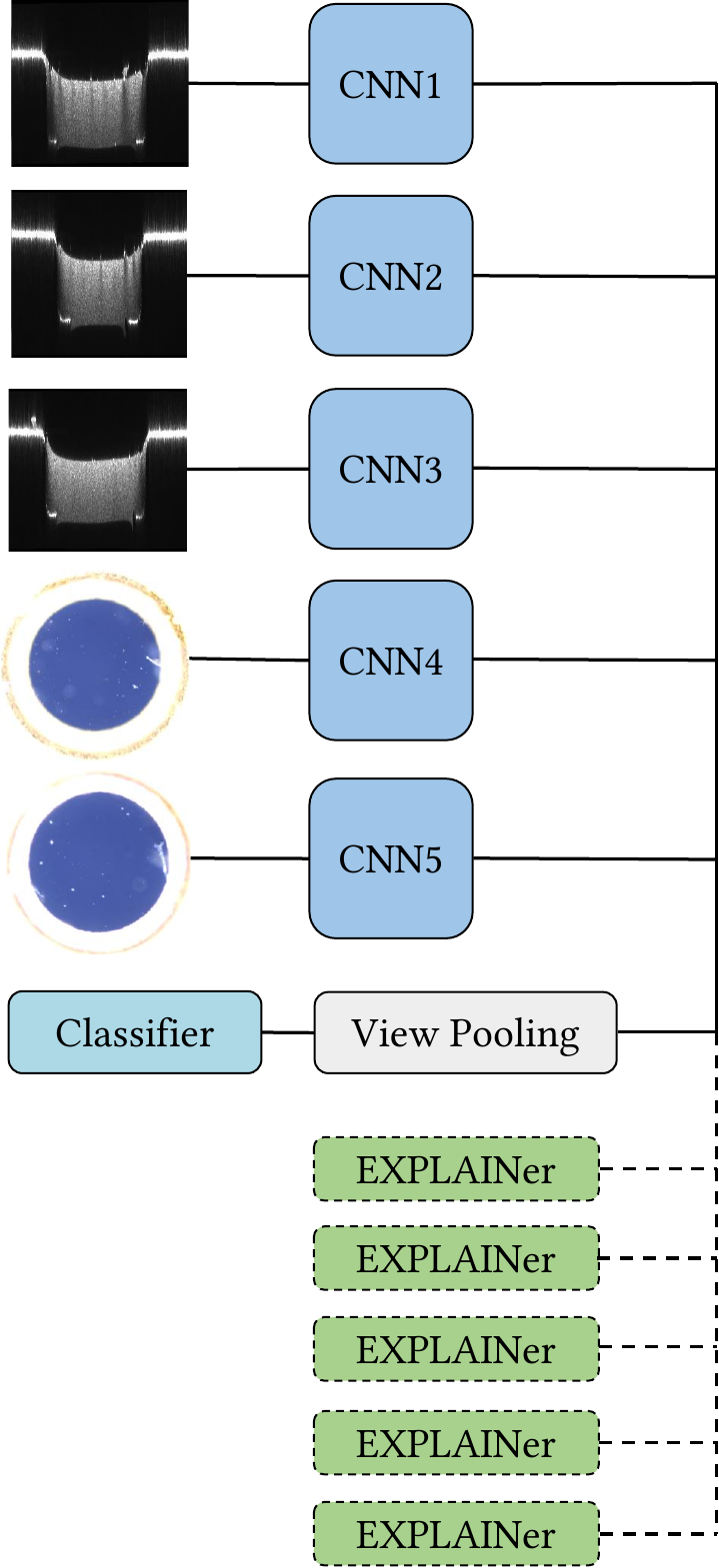}}\hspace{1em}%
  \vspace{-0.8cm}
  \caption{Demonstrating the Multi-View architectures correspondingly to \autoref{fig:multi_view_architectures} on a real-life physical problem. In this case, there are 5 different views and 2 visually different sub-groups in the foam quality assessment for the HEDP experiments domain. (a) refers to all different views as visually similar, which is wrong. (b) does the correct splitting between the sub-groups, (c) can be correct to some degree and (d) has redundant feature extractors assuming each view is completely different than the other. EXPLAINer is connected to each feature extractor to explain its decision.}
  \label{fig:multi_view_demonstrate}
\end{figure*}

%% file: figures_5.tex
\begin{table*}
    \centering
    \begin{tabular} {| M{0.5cm} | M{2.9cm} | M{2.9cm} | M{2.9cm} | M{2.9cm} |}
        \hline
        & CSV & SSG & PSG & CDV \\  
        \hline
        \multicolumn{1}{|c|}{\rotatebox[origin=c]{90}{\centering Loss Trend}}  & 
        \includegraphics[height=2.0cm,width=3cm]{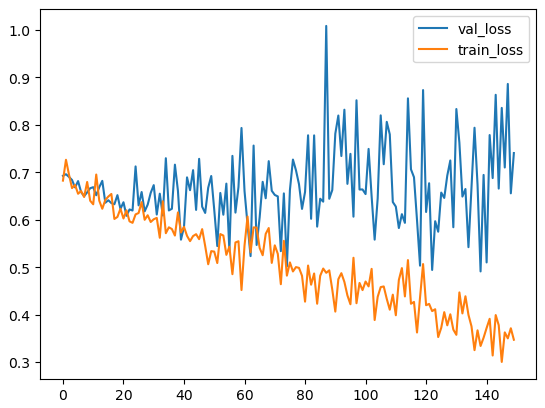} &
        \cellcolor[HTML]{D5DBDB} \includegraphics[height=2.0cm,width=3cm]{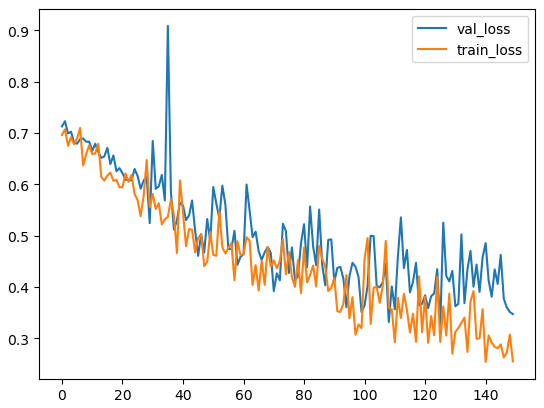} &
        \includegraphics[height=2.0cm,width=3cm]{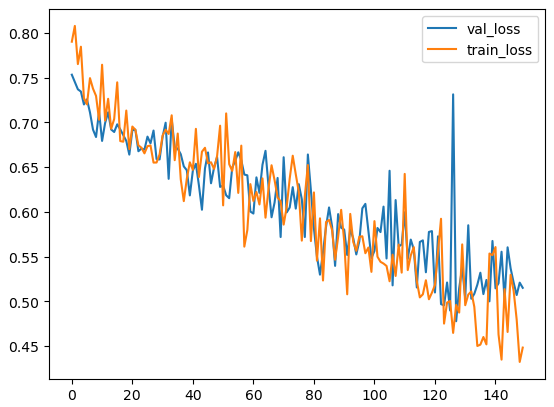} &
        \includegraphics[height=2.0cm,width=3cm]{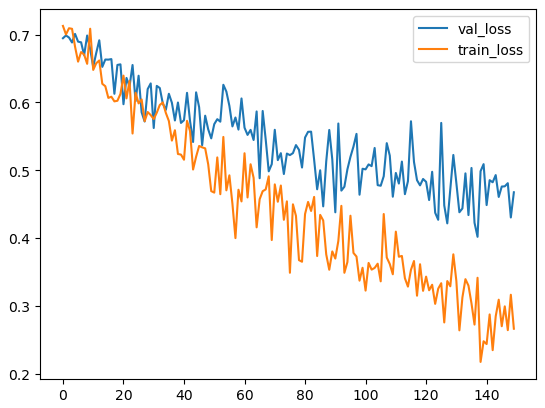}
        \\ \hline

        \multicolumn{1}{|c|}{\rotatebox[origin=c]{90}{\centering Acc Trend}} & 
        \includegraphics[height=2.0cm,width=3cm]{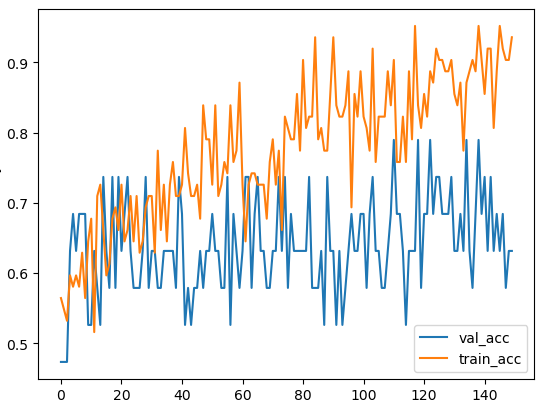} &
        \cellcolor[HTML]{D5DBDB} \includegraphics[height=2.0cm,width=3cm]{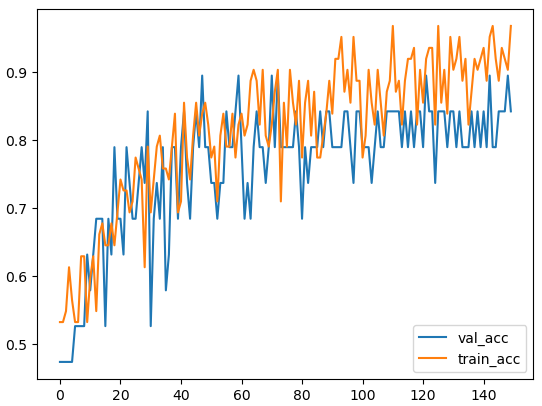} &
        \includegraphics[height=2.0cm,width=3cm]{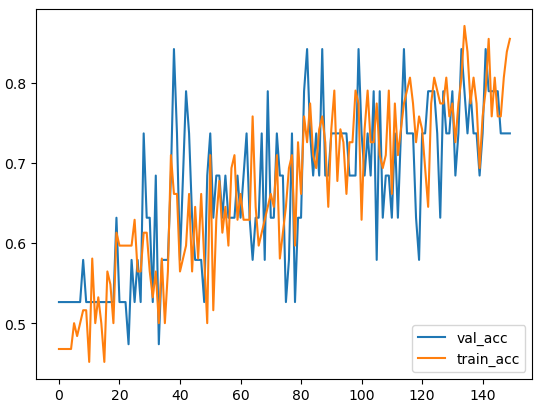} &
        \includegraphics[height=2.0cm,width=3cm]{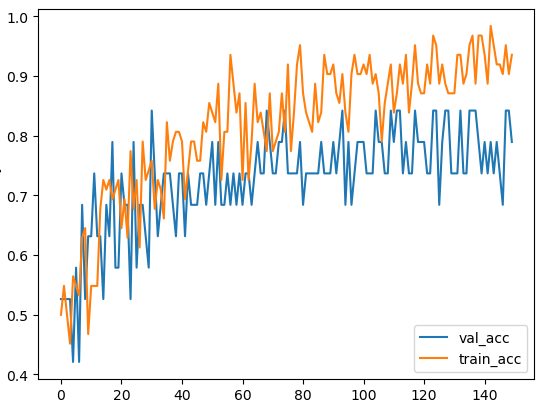}
        \\ \hline 
        Acc & 78 & \textbf{84} & 78 & 84 \\ \hline
        AUC & 83 & \textbf{93} & 81 & 84  \\  \hline
    \end{tabular}
    \caption{Learning curves (accuracy and loss as functions of epochs) of all models. The best model's cells are shadowed. These empirical results demonstrate that SSG is the best architecture for the given use case. While CSV and CDV develop overfit in epoch 60, SSG and PSG continue to converge. SSG architecture confirmed to result in the best performance with an accuracy of 84\% and AUC of 93\% compared to 78\% and 81\% by PSG.}
    \label{table:learning_curve}
\end{table*}

%% file: figures_4.tex
\begin{figure*}[!ht]
  \centering
  \captionsetup[subfigure]{labelformat=empty}
  \subcaptionbox{\label{limefig:a}}{\includegraphics[width=0.7in]{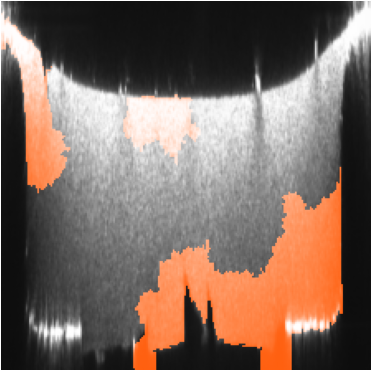}}\hspace{1em}%
  \subcaptionbox{\label{limefig:b}}{\includegraphics[width=0.7in]{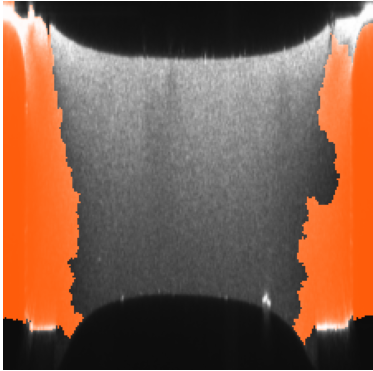}}\hspace{1em}%
  \subcaptionbox{\label{limefig:c}}{\includegraphics[width=0.7in]{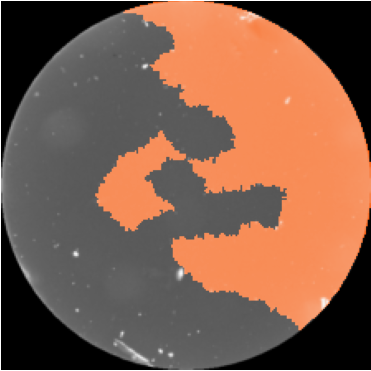}}\hspace{1em}%
  \subcaptionbox{\label{limefig:d}}{\includegraphics[width=0.7in]{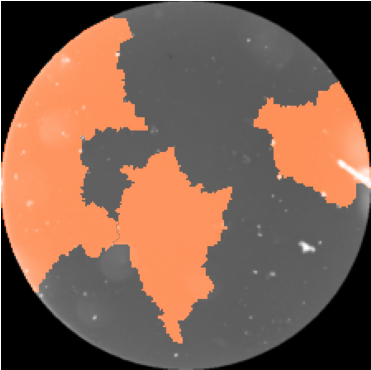}}
  \vspace{-0.7cm}
  \caption{Multi-View local explainability with LIME. Areas of interest are marked with orange in each view. Cracks in the profiles can be watched, indicating a contribution towards defective foam decision. On the other hand, empty areas are marked in the top-bottom views, contributing towards a normal decision.}
  \label{fig:LIME_attention}
\end{figure*}

\begin{figure*}[!ht]
  \centering
  \captionsetup[subfigure]{labelformat=empty}
  \subcaptionbox{\label{limefig:a}}{\includegraphics[width=1.6in]{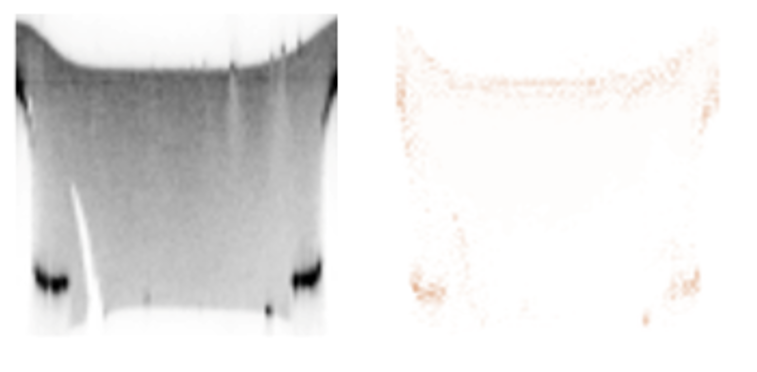}}\hspace{1em}%
  \subcaptionbox{\label{limefig:b}}{\includegraphics[width=1.6in]{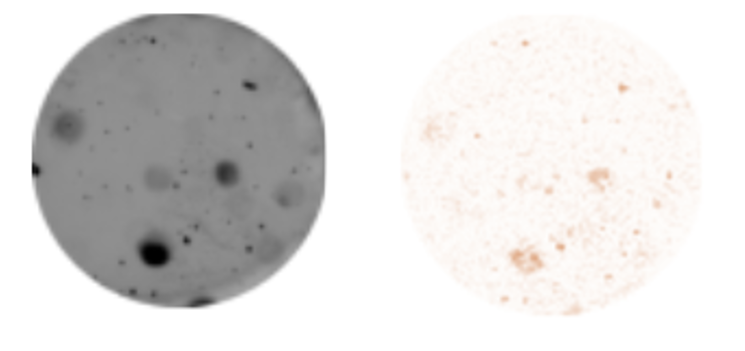}}\hspace{1em}%
  \vspace{-0.5cm}
  \caption{Multi-View global explainability of a profile and top-bottom views with SHAP. Pixels that contribute to the prediction are marked with orange.}
  \label{fig:SHAP}
  \vspace{-0.3cm}
\end{figure*}